\documentclass[sigconf, screen]{acmart}

\usepackage{titlesec}
\usepackage{float}
\usepackage{array}
\usepackage{booktabs}
\usepackage{multirow}
\usepackage{graphicx}
\usepackage{animate}
\usepackage{makecell}
\usepackage{subcaption}

\usepackage{pifont}

\newcommand{\cmark}{\ding{51}}  
\newcommand{\xmark}{\ding{55}}  

\titleformat{\paragraph}
  [block] 
  {\normalfont\normalsize\bfseries} 
  {\theparagraph}{1em}{} 
\titlespacing*{\paragraph}
  {0pt}{1ex plus .2ex minus .2ex}{1ex plus .2ex}

\titleformat{\subsubsection}
  {\normalfont\normalsize\bfseries}
  {\thesubsubsection}{1em}{}

\setcopyright{acmlicensed}
\copyrightyear{2025}
\acmYear{2025}
\acmDOI{XXXXXXX.XXXXXXX}
\acmConference[Conference acronym -]{Make sure to enter the correct
  conference title from your rights confirmation email}{-}{-}

\acmISBN{978-1-4503-XXXX-X/2018/06}

\begin{document}

\title{BrokenVideos: A Benchmark Dataset for Fine-Grained Artifact Localization in AI-Generated Videos}

\author{Jiahao Lin}
\authornote{Equal Contribution}
\affiliation{
  \institution{Fudan University}
  \city{Shanghai}
  \country{China}
}
\email{23210240235@m.fudan.edu.cn}

\author{Weixun Peng\scriptsize$^{*}$}
\affiliation{
  \institution{Shenzhen University}
  \city{Shenzhen}
  \country{China}
}
\email{pengweixuan2023@email.szu.edu.cn}

\author{Bojia Zi}
\affiliation{
  \institution{The Chinese University of Hong Kong}
  \city{Hong Kong SAR}
  \country{China}
}
\email{bjzi@se.cuhk.edu.hk}

\author{Yifeng Gao}
\affiliation{
  \institution{Fudan University}
  \city{Shanghai}
  \country{China}
}
\email{yifenggao23@m.fudan.edu.cn}

\author{Xianbiao Qi}
\affiliation{
  \institution{IntelliFusion Inc.}
  \city{Shenzhen}
  \country{China}
}
\email{qixianbiao@gmail.com}

\author{Xingjun Ma}
\affiliation{
  \institution{Fudan University}
  \city{Shanghai}
  \country{China}
}
\email{xingjunma@fudan.edu.cn}

\author{Yu-Gang Jiang}
\affiliation{
  \institution{Fudan University}
  \city{Shanghai}
  \country{China}
}
\email{ygj@fudan.edu.cn}

\begin{abstract}
The field of video generation has witnessed remarkable advances in recent years, driven by innovations in deep generative models. Nevertheless, the fidelity of AI-generated videos remains far from perfect, with synthesized content frequently exhibiting visual artifacts, such as temporally inconsistent motion, physically implausible trajectories, unnatural object deformations, and local blurring, that undermine realism and user trust. Precise detection and spatial localization of these artifacts are of critical importance: not only are they essential for automatic quality control pipelines that improves user experience, but they also provide actionable diagnostic signals for researchers and practitioners to guide model development and evaluation.
Despite its significance, the research community currently lacks a comprehensive benchmark tailored for artifact localization in AI-generated videos. Existing datasets either focus solely on detection at the video or frame level, or lack fine-grained spatial annotations necessary for developing and benchmarking localization methods. To fill this gap, we present BrokenVideos, a benchmark dataset comprising $\sim$3,254 AI-generated videos with carefully-annotated, pixel-level masks indicating regions of visual corruption. Each annotation is the result of careful human inspection, ensuring high-quality ground truth for artifact localization tasks.
We demonstrate that training existing video artifact detection models and multi-modal large language models (MLLMs) on BrokenVideos substantially enhances their ability to localize corrupted regions within generated content. Through extensive experiments and cross-model evaluations, we show that BrokenVideos provides a critical foundation for both benchmarking and advancing artifact localization research. We hope our dataset can catalyze further innovation in both video generation and its quality assurance. The dataset is available at: \url{https://broken-video-detection-datetsets.github.io/Broken-Video-Detection-Datasets.github.io/}.
\end{abstract}



\maketitle

\begin{figure*}[!htp]
\centering
\animategraphics[autoplay,loop,width=0.875\linewidth]{10}{image/teasers/test_}{0}{31}
\vspace{-0.1in}
\caption{Visualization of BrokenVideos. \emph{Best viewed using Acrobat Reader. Click on the images to play the animation clips.}}
\end{figure*}

\section{INTRODUCTION}

Recent breakthroughs in diffusion-based video generation have elevated text-to-video synthesis to unprecedented levels, yielding outputs with remarkable coherence, temporal consistency, and visual fidelity. Closed-source models such as Gen3 ~\cite{gen1_esser2023structure} and SORA ~\cite{sora} have set new standards for prompt fidelity and aesthetic quality. Meanwhile, open-source approaches like CogVideoX ~\cite{cogvideox_yang2024cogvideox}, HunyuanVideo ~\cite{hunyuanvideosystematicframeworklargekong2025}, and Wan 2.1 ~\cite{wan2025wanopenadvancedlargescale} are rapidly narrowing the performance gap through innovative architectures and large-scale training. The push toward increasingly large and sophisticated models is evident: Genmo AI’s Mocchi ~\cite{mochi_1}, with 10 billion parameters, demonstrates exceptional motion fidelity, while STEP-Video ~\cite{ma2025step} employs a 30-billion-parameter hybrid architecture to generate long-form, cross-lingual videos.

Despite these advances, the fidelity of AI-generated videos remains far from perfect. Even the most advanced models frequently produce \textbf{visual artifacts}, such as temporally inconsistent motion, unnatural object deformations, and local blurring—that disrupt the illusion of realism and diminish user trust. In practical applications, these “broken videos” compromise the reliability and acceptance of generative models, necessitating post-hoc quality control pipelines to automatically select or filter outputs.

However, existing research has largely overlooked the problem of artifact localization in AI-generated videos. The closest related task in the community is forensic classification, which aims to determine whether a given video is real or synthetic—often by implicitly checking for the presence of artifacts. Frameworks such as AIGVDet ~\cite{vahdati2024deepfakeimagesdetecting}, DeCoF ~\cite{bai2024aigenerated}, and DIVID ~\cite{liu2024turnsimrealrobust} have achieved strong performance in this binary classification setting. However, their outputs are limited to a single yes/no label, offering limited insight into which regions of a video contain artifacts. This lack of spatial specificity fundamentally limits their utility for fine-grained artifact localization and diagnostic analysis.

To address this gap, we introduce \textbf{BrokenVideos}, a fine-grained benchmark dataset comprising 3,254 AI-generated videos, each annotated with pixel-level masks for artifact detection. Our dataset features substantial diversity, as the videos are sourced from a wide range of state-of-the-art video generation methods. Leveraging BrokenVideos, we successfully trained multiple segmentation-based detectors capable of accurately localizing corrupted regions within generated content.

The key contributions of this work are as follows:

\begin{itemize}
\item We introduce \textbf{BrokenVideos}, a benchmark dataset for video artifact localization, comprising 3,254 AI-generated videos selected from diverse generation methods.

\item Each video in BrokenVideos is annotated with detailed, pixel-level segmentation masks for artifacts, using the SAM2 interactive annotation tool.

\item We fine-tune multiple segmentation models on BrokenVideos, achieving state-of-the-art performance in artifact localization. This demonstrates the usefulness of our dataset for both detector training and performance evaluation.
\end{itemize}

\section{RELATED WORK}
\subsection{Video Generation}
Video generative models have witnessed rapid advancements in recent years. Early approaches, such as Make-A-Video ~\cite{makeavideo_singer2022make} and ModelScope ~\cite{modelscope_wang2023modelscope}, were among the first to synthesize dynamic animations from textual prompts. Building on these foundations, open-source methods like AnimateDiff ~\cite{animatediff_guo2023animatediff} and VideoCrafter ~\cite{videocrafter2_chen2024videocrafter2} adopted CNN-based UNet architectures to generate videos, achieving notable performance improvements.

A significant milestone in the field is VideoPoet ~\cite{videopoet_kondratyuk2023videopoet}, which introduced an auto-regressive framework for video generation. More recently, research has shifted toward leveraging DiT-based models to further enhance video synthesis quality. For instance, CogVideoX ~\cite{cogvideox_yang2024cogvideox} utilizes a two-stream expert architecture to process text and video information separately, while HunyuanVideo ~\cite{hunyuanvideosystematicframeworklargekong2025} incorporates both two-stream and fusion experts, enabling joint learning and fusion of textual and visual data through attention mechanisms. Additionally, Wan 2.1 ~\cite{wan2025wanopenadvancedlargescale} represents a significant advancement by employing self-attention for modeling video content and cross-attention to integrate textual information.

While open-source models have made substantial progress, closed-source systems have pushed the boundaries even further. Notable examples include SORA ~\cite{sora}, Kling ~\cite{keling}, and Gen3 ~\cite{gen3}, which have attracted widespread attention for their ability to transform creative prompts into highly realistic and coherent videos.

Despite remarkable progress in both open-source and closed-source video generation, current models still exhibit notable imperfections, often manifesting as various visual artifacts in the generated content.

\subsection{Video Object Segmentation}
Fine-grained artifact localization can be framed as a video object segmentation (VOS) task, aiming to accurately identify and localize broken objects/regions within AI-generated videos.
VOS is a longstanding and fundamental problem in computer vision, focused on accurately delineating and tracking semantically meaningful objects across video sequences. Leading VOS models, such as OSVOS ~\cite{caelles2017oneshotvideoobjectsegmentation}, SAMWISE ~\cite{cuttano2025samwiseinfusingwisdomsam2}, and MatAnyone ~\cite{yang2025matanyonestablevideomatting}, leverage spatiotemporal coherence and hierarchical feature representations to achieve robust segmentation of foreground entities under challenging conditions. These approaches have achieved state-of-the-art performance on established benchmarks such as DAVIS and YouTube-VOS, where the primary objective is to segment objects like humans, animals, and vehicles under diverse motion, occlusion, and appearance variations.
Artifact localization in AI-generated videos, however, presents a new challenge. Unlike natural objects, artifacts are often irregular, amorphous, and lack clear semantic meaning, making them difficult to identify for models trained solely on natural scene segmentation. Nevertheless, both artifact localization and VOS require high spatial precision and temporal consistency.


\begin{figure*}[!ht]
    \centering
    \includegraphics[width=0.9\linewidth]{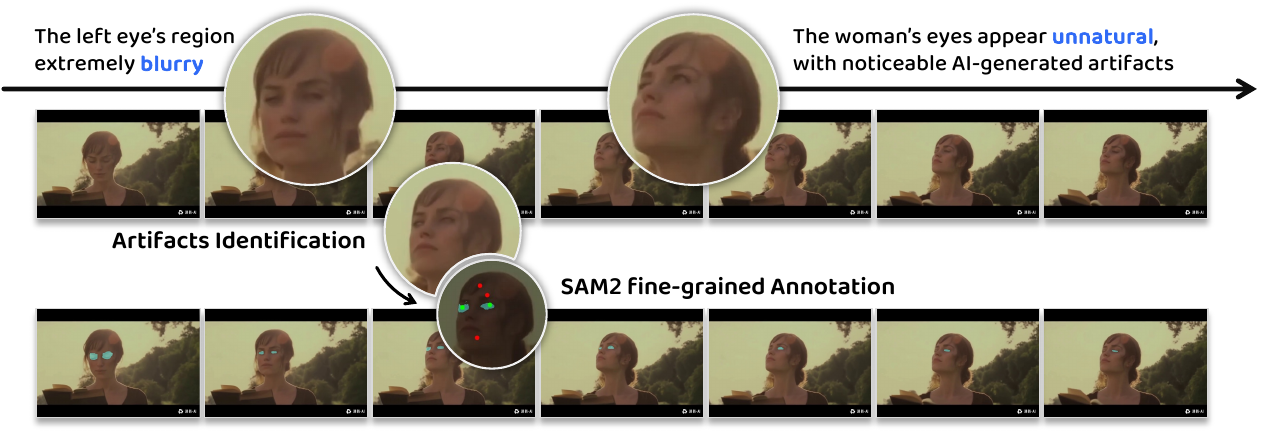}
    \caption{Visualization of our artifact annotation strategy.}
    \label{fig:pipeline}
\end{figure*}

\subsection{Video Artifact Detection}

\textbf{Detection Methods}\;
Contemporary approaches to video artifact detection can be broadly classified into two main paradigms. The first centers on detecting specific artifact types using dedicated algorithms. For example, CAMBI ~\cite{tandon2021cambicontrastawaremultiscalebanding} and BBAND ~\cite{tu2020bbandindexnoreferencebanding} are specialized in identifying banding artifacts, while EFENet ~\cite{9447919} and MLDBD ~\cite{10054073} focus on spatial blur. Other techniques, such as ~\cite{Wolf2008ANR}, are tailored for frame drop detection. More comprehensive frameworks like VIDMAP ~\cite{8585129} deploy multiple independent models to quantify a range of artifact categories. However, a fundamental limitation of these approaches is the implicit assumption that artifacts occur in isolation. In reality, AI-generated videos frequently exhibit complex combinations of co-occurring and interacting artifacts, which can confound single-artifact detectors and reduce their effectiveness in practice.
The second paradigm employs video quality assessment (VQA) metrics to infer the presence of artifacts by thresholding objective quality scores. For instance, the method proposed in ~\cite{Wu_2023} is capable of detecting several common artifacts introduced during video acquisition and delivery. Similarly, some studies utilize static thresholds on established quality metrics to identify compression-related degradations ~\cite{8585129}. While VQA-based methods offer a degree of generality, they typically lack the spatial precision and sensitivity needed for accurate artifact localization. These methods are less effective than purpose-built artifact detectors, particularly in complex or mixed-artifact scenarios.

\textbf{Detection Datasets} \;
Progress in artifact detection research is constrained by the availability of high-quality, diverse datasets. Currently, only two public databases, Maxwell ~\cite{Wu_2023} and BVI-Artifact ~\cite{feng2024bviartefactartefactdetectionbenchmark}, provide video sequences annotated with multiple artifact types. Maxwell consists of 4,543 user-generated videos spanning a wide range of spatial resolutions, each potentially exhibiting up to eight artifact categories. In contrast, BVI-Artifact includes 480 professionally produced HD and UHD videos, annotated for up to six artifact types per sequence. While both datasets have propelled the field forward, they are limited in scale and diversity, and—critically—do not address artifacts unique to AI-generated content. This gap highlights the urgent need for a comprehensive, fine-grained dataset that enables rigorous evaluation and development of artifact detection and localization methods tailored to the novel challenges posed by modern video generation models.

\section{BrokenVideos Dataset}
In this section, we present \textbf{BrokenVideos}, a benchmark dataset curated for localizing artifacts in AI-generated videos. BrokenVideos is built upon a rigorous, operational definition of video artifacts and results from a comprehensive pipeline including prompt engineering, large-scale video generation, and region-level, pixel-accurate annotation. The dataset comprises diverse video content, each annotated with fine-grained artifact labels, enabling robust training and evaluation of artifact detection and localization models.

\subsection{Motivation}
Our motivation for constructing the BrokenVideos dataset stems from a comprehensive and systematic review of existing benchmarks and evaluation protocols for AIGC-generated video quality assessment, which reveals several critical gaps in the current landscape. First, most available datasets either lack precise, region-level annotations of artifacts or focus solely on global perceptual quality, failing to capture the spatial and temporal localization of common artifacts that frequently occur in state-of-the-art generative models. Second, there is no unified, operational definition of \emph{video artifacts}, resulting in inconsistencies in annotation and evaluation across different studies. Third, existing datasets often lack diversity in scenes, artifact types, and generation models, limiting their effectiveness for developing and benchmarking robust artifact localization methods.

These observations highlight the urgent need for a dedicated dataset that addresses these shortcomings. Intuitively, an ideal dataset for artifact localization should meet several key criteria: (1) a rigorous and reproducible definition of video artifacts; (2) fine-grained, pixel-level annotations that support both detection and localization tasks; (3) broad diversity in scenes and source generation models to ensure generalizability; and (4) sufficient scale and annotation quality to enable the training of more powerful video generation models and statistically meaningful video quality evaluation.
Our proposed \textbf{BrokenVideos} dataset is explicitly designed to fulfill these requirements. By providing a large-scale, richly annotated, and conceptually coherent benchmark, BrokenVideos aims to advance the development, evaluation, and understanding of artifact localization methods for AI-generated videos.

\begin{table*}[!htp]
\centering
\caption{Summary of artifact (sub)categories.}
\renewcommand{\arraystretch}{0.9}
\begin{tabular}{
    >{\raggedright\arraybackslash}m{2.8cm} |
    >{\raggedright\arraybackslash}m{3.3cm}   |
    >{\raggedright\arraybackslash}m{4cm}   |
    >{\raggedright\arraybackslash}m{5cm}
}
\toprule
\textbf{Category} & 
\textbf{Subcategory} & 
\textbf{Description} & 
\textbf{Example} \\
\midrule
\multirow{2}{*}{Technical Artifact} 
    & Visual breakdown & fragmented, warped, or disjointed content & "a face melting into abstract shapes" \\
    \cline{2-4}
    & Motion anomalies & unnatural transitions or temporal jumps & "legs bending backward while walking" \\
\midrule
\multirow{3}{*}{Semantic Artifact}
    & Symbol confusion & culturally nonsensical or meaningless text & "" \\
    \cline{2-4}
    & Logical disjunction & causally incoherent scene transitions & "sudden disappearance or identity swap" \\
    \cline{2-4}
    & Emotional misalignment & actions and environment express conflicting emotions & "" \\
\midrule
\multirow{2}{*}{Cognitive Artifact}
    & Uncanny valley effect & unsettling abnormalities in humanoid figures & "asynchronous eye movement" \\
    \cline{2-4}
    & Perceptual overload & high-density meaningless information & "fragmented objects filling the entire frame" \\
\midrule
\multirow{2}{*}{Consistency Artifact}
    & Style inconsistency & abrupt or jarring transitions between visual styles & "realism turning to cartoon" \\
    \cline{2-4}
    & Attribute shift & object properties change inconsistently & "clothing color or age fluctuating across frames" \\
\midrule
\multirow{2}{*}{Instructional Artifact}
    & Prompt misalignment & output deviates significantly from intended description & "generating a 'volcanic eruption' for 'peaceful beach'" \\
    \cline{2-4}
    & Omission of key details & Omission of key details & "missing glasses on a character explicitly described as wearing them" \\
\bottomrule
\end{tabular}
\label{tab:category}
\end{table*}

\subsection{Dataset Construction}

Given the practical constraints of human and computational resources, we adopt a hybrid dataset construction strategy. In addition to a subset of in-house generated videos, we leverage the existing benchmarking dataset~\cite{zeng2024dawnvideogenerationpreliminary}. Broken videos from both sources were systematically identified and annotated with fine-grained masks, thereby ensuring both diversity and annotation quality while minimizing computational costs.

\subsubsection{Artifact Categorization and Annotation}
To establish a robust and reproducible framework for video artifact annotation, we devised a comprehensive taxonomy informed by systematic manual examination of a large corpus of generated samples, further enriched by the analytical capabilities of state-of-the-art large language models e.g., GPT-4o~\cite{openai2023gpt4}. This iterative process of synthesis and validation ensures that our taxonomy is not only empirically grounded but also sufficiently general to encompass the diverse failure modes encountered in contemporary generative video content.

As summarized in Table ~\ref{tab:category}, we delineate five main categories of artifacts: \emph{Technical Artifacts}, \emph{Semantic Artifacts}, \emph{Cognitive Artifacts}, \emph{Consistency Artifacts}, and \emph{Instructional Artifacts}. Each category encapsulates a distinct dimension of failure, ranging from low-level visual or motion aberrations to higher-order inconsistencies in semantics, cognition, style, and adherence to prompts.

A salient and challenging aspect of our annotation protocol is the frequent co-occurrence of multiple artifact types within a single video. In practice, artifact manifestations often overlap both spatially and temporally, necessitating a multi-label, multi-instance annotation strategy. Annotators are required not only to precisely localize artifact regions at the frame level, but also to accurately assign them to the appropriate categories within our taxonomy. This dual requirement imposes substantial cognitive and operational demands, resulting in a markedly increased annotation workload and complexity. Our protocol is thus designed to rigorously capture the nuanced and multifaceted nature of artifact emergence in AI-generated videos, ultimately yielding a richly annotated dataset that supports advanced analysis and model development.

To efficiently and accurately annotate artifact regions in small-scale video collections, we develop a semi-automatic graphical user interface tool based on Segment Anything Model 2 (SAM2)~\cite{sam2_ravi2024sam2}, referred to as SAM2-GUI, to accelerate the frame-level mask annotation process. The tool allows sequential loading and browsing of video frames. Users can interactively generate candidate masks by clicking positive and negative prompts, with real-time visual editing. Importantly, the system supports annotation of multiple artifact regions within a single frame, generating multi-instance masks to capture cases where multiple regions are corrupted simultaneously.

Once initial annotation is completed, masks can be submitted to a tracking module for temporal propagation. Leveraging the video tracking capabilities of SAM2, the tool maintains spatial consistency across frames while reducing manual effort. The propagated results are visualized for user inspection and frame-by-frame refinement, ensuring high-quality annotations. Finally, all masks are saved as frame-level PNGs and can be exported as visualization videos for downstream validation and model training.


\subsection{Dataset Statistics}
A comprehensive understanding of the dataset’s scale, diversity, and artifact characteristics is essential for benchmarking and advancing artifact localization in AI-generated videos. We present a systematic analysis of our dataset from global properties to core innovations.

\textbf{Dataset Scale and General Properties}\; 
Our dataset comprises \textbf{3,254} high-resolution AI-generated videos, spanning a wide range of resolutions, from 416*624 to 1760*1152. Each video has been meticulously annotated, resulting in a total of $\sim$ \textbf{336,000} high-quality manual artifact region annotations. This unprecedented scale provides a robust foundation for both training and evaluating artifact localization methods.

\begin{figure}[!ht]
    \centering
    \includegraphics[width=0.9\linewidth]{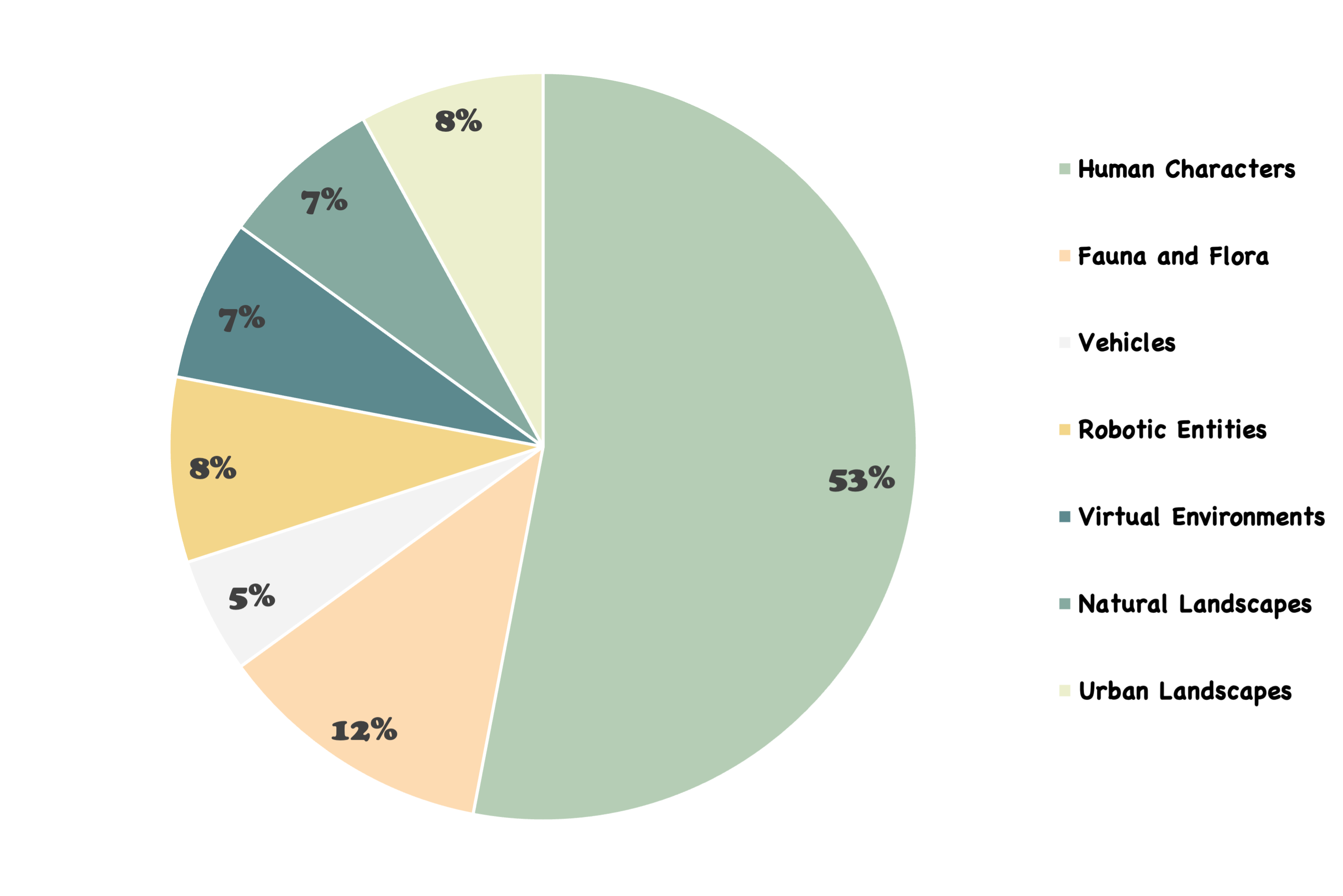}
    \caption{Topic distribution of videos in BrokenVideos.}
    \label{fig:category}
    \vspace{-0.1in}
\end{figure}

\textbf{Scene Diversity}\; 
To ensure comprehensive coverage, our dataset encompasses a diverse range of scene categories and generative model sources. Specifically, it includes a variety of semantic scene types such as \textbf{Human Characters} (53\%), \textbf{Fauna and Flora} (12\%), \textbf{Robotic Entities} (8\%), \textbf{Urban Landscapes} (8\%), \textbf{Virtual Environments} (7\%), \textbf{Natural Landscapes} (7\%), and \textbf{Vehicles} (5\%). A detailed distribution of these categories is illustrated in Figure~\ref{fig:category}.

\textbf{Artifact Region Distribution}\;
A key innovation of our dataset is the detailed annotation and statistical profiling of artifact (defect) regions within each video. Analysis reveals that \textbf{80\%} of videos contain \textbf{1–2} artifact regions, \textbf{17\%} contain \textbf{3–6} regions, and \textbf{3\%} have \textbf{7–10} regions. This distribution not only reflects the natural sparsity of severe artifacts in modern generative models, but also provides sufficient challenging cases for robust benchmarking. The granularity of these annotations enables the evaluation of localization algorithms in both typical and complex, multi-defect scenarios.

\begin{figure*}[!htp]
\centering
\animategraphics[autoplay,loop,width=0.8\linewidth]{10}{image/comparison/test_}{0}{31}
\caption{Comparison of various baseline artifact localization models and models trained on our BrokenVideos dataset. \emph{Best viewed using Acrobat Reader. Click on the images to play the animation clips.}}
\end{figure*}






\begin{table*}[!htp]
\centering
\caption{Evaluation results of various localization models on our \textbf{BrokenVideos} dataset. Best results are highlighted in \textbf{bold}.}
\label{tab:results}
\begin{tabular}{c|c|ccc|ccc|ccc}
\toprule
\multirow{2}{*}{\textbf{Method}}  & \multirow{2}{*}{\makecell{\textbf{Finetuned on} \\ \textbf{BrokenVideos}}}& \multicolumn{3}{c|}{\textbf{Broken Videos}} &\multicolumn{3}{c|}{\textbf{Normal Videos}} &\multicolumn{3}{c}{\textbf{All Videos}}\\ \cline{3-11}
& & \textbf{J} & \textbf{F} & \textbf{J\&F} & \textbf{J} & \textbf{F} & \textbf{J\&F} & \textbf{J} & \textbf{F} & \textbf{J\&F} \\
\hline
\multirow{2}{*}{GLUS} & \xmark &0.1312 & 0.1091 & 0.1202 & 0.2727 & 0.0469 & 0.1598  & 0.2020 & 0.0780 & 0.1400\\
\cline{2-11}
&  \cmark & \textbf{0.6700} & \textbf{0.6074} & \textbf{0.6387} & \textbf{0.4795} & \textbf{0.4768} & \textbf{0.4782} & \textbf{0.5748} & \textbf{0.5421} & \textbf{0.5585}\\
\hline
\multirow{2}{*}{SAMWISE} & \xmark & 0.0594 & 0.0431 & 0.0513 & 0.0898 & 0.0196 & 0.0547 & 0.0746 & 0.0320 & 0.0530\\
\cline{2-11}
& \cmark & \textbf{0.6523} & \textbf{0.4284} & \textbf{0.5403} & \textbf{0.5116} & \textbf{0.3436} & \textbf{0.4276} & \textbf{0.5820} & \textbf{0.3860} & \textbf{0.4840}\\ \hline 
\multirow{2}{*}{VideoLISA} & \xmark & 0.2936 & 0.1830 & 0.2383 & 0.3042 & 0.1522 & 0.2282 & 0.2989 & 0.1676 & 0.2333\\
\cline{2-11}
 & \cmark & \textbf{0.6679} & \textbf{0.5924} & \textbf{0.6301} & \textbf{0.4974} & \textbf{0.4960} & \textbf{0.4967} & \textbf{0.5827} & \textbf{0.5442} & \textbf{0.5634}\\ 
\bottomrule
\end{tabular}
\end{table*}

\section{Fine-tuning Artifact Localization Models}

\subsection{Training and Test Datasets}

To train the localization model, we use 3,091 broken videos in our dataset with manually annotated masks. Besides, we also use 748 clean videos, with masks fully labeled as normal since they contain no artifacts, in order to prevent model overfitting on the artifact videos and encourage it has ability to recognize the normal videos(the videos without artifacts). For those localization model, which requires the prompt guidance, the artifact regions are paired with the prompt: "\textit{There’s something uncanny in the video. The video suffers from visual artifacts.}" The normal regions with the prompt: "\textit{This video has no artifacts and everything is normal.}" For validation, we use a total of 300 videos, consisting of 150 broken and 150 normal videos. We evaluate the performance of our trained localization model using the region similarity $\mathcal{J}$, contour accuracy $\mathcal{F}$, and their average $\mathcal{J} \& \mathcal{F}$, which are commonly adopted in video object segmentation tasks to measure mask quality.

\subsection{Model Fine-Tuning}

Here, we selected 3 popular MLLM-based localization models, including SAMWISE~\cite{cuttano2024samwise}, GLUS~\cite{lin2025glus} and VideoLISA~\cite{videolisa}. We finetuned these models on our dataset.  We fine-tuned these models on 8 A800 GPUs, each GPU has 80GB memory.
For SAMWISE, we fine-tuned it for 8 epochs with constant learning rate of 1e-5 and global batch size of 16, using the AdamW optimizer~\cite{adamw_loshchilov2017decoupled}. Following the default settings, we do not use weight decay. 
For GLUS, we finetuned 3000 steps with the default learning rate of 3e-4, weight decay is 0. We use warmup decay learning rate scheduler with 100 warmup steps. Each step corresponds to a batch size of 5 per device and 2 gradient accumulation steps. 
For VideoLISA, we fine-tuned it for 2000 steps using the AdamW optimizer with the learning rate and weight decay set to 3e-4 and 0, respectively. Flowing their configuration, we also adopt WarmupDecayLR as the learning rate scheduler, with the warmup steps are set to 100.

\subsection{Performance Evaluation}

To evaluate the effectiveness of \textbf{BrokenVideos}, we conduct experiments using three representative video localization models: GLUS~\cite{lin2025glus}, SAMWISE~\cite{cuttano2024samwise}, and VideoLISA~\cite{videolisa}. We test each model both before and after fine-tuning on BrokenVideos. As shown in Table \ref{tab:results}, fine-tuning leads to significant performance improvements across all metrics, including J, F, and J\&F scores, particularly on videos containing artifacts. For example, GLUS achieves a J\&F score of 0.6387 after fine-tuning, compared to only 0.1202 without it. SAMWISE and VideoLISA show similar trends, improving by 0.4890 and 0.3918 respectively. This confirms that BrokenVideos effectively enables models to learn meaningful representations of artifact regions, which are difficult to capture using models trained only on clean video data. Notably, we also observe considerable gains on normal videos, suggesting that the learned features generalize beyond artifact-heavy scenarios. For instance, the J\&F score of VideoLISA on normal videos increases from 0.2282 to 0.4967 after fine-tuning. These improvements are consistent across region accuracy (J) and boundary accuracy (F), indicating better overall segmentation performance, including sharper and more coherent object boundaries. The consistent performance gains across artifact, normal, and mixed video sets demonstrate that \textbf{BrokenVideos} serves as an effective supervisory signal for both artifact detection and general video localization. The dataset significantly enhances model robustness and accuracy, making it a valuable resource for advancing video understanding in both clean and degraded conditions.

\section{Conclusion}
In this paper, we present \textbf{BrokenVideos}, a carefully curated dataset comprising 3,254 AI-generated videos from multiple models, each paired with human-verified, pixel-level masks precisely identifying visual artifacts. Every annotation was meticulously reviewed to ensure the high fidelity essential for artifact localization research. Extensive experiments demonstrate that training state-of-the-art video MLLMs on BrokenVideos significantly enhances their ability to localize broken regions in synthetic content. Cross-model evaluations further confirm that BrokenVideos serves as a critical, realistic benchmark for artifact localization, paving the way for more robust and generalizable solutions. We hope this dataset will enable the research community to achieve new advances in video generation and quality assurance.

\bibliographystyle{ACM-Reference-Format}
\bibliography{abbrev}

\end{document}